\documentclass[runningheads]{llncs}
\usepackage{graphicx}

\usepackage{tikz}
\usepackage{comment}
\usepackage{amsmath,amssymb} 
\usepackage{color}

\usepackage[accsupp]{axessibility}  
\usepackage{amssymb}
\usepackage{amsmath}
\usepackage{graphicx}
\usepackage{subfigure}
\usepackage{enumitem}
\usepackage{multirow}
\usepackage{color}
\usepackage{booktabs}
\usepackage{algorithmic,algorithm}

\usepackage[pagebackref,breaklinks,colorlinks,bookmarks=false]{hyperref}
\usepackage[capitalize]{cleveref}

\begin{document}
\pagestyle{headings}
\mainmatter
\def\ECCVSubNumber{1589}  

\title{CGUA: Context-Guided and Unpaired-Assisted Weakly Supervised Person Search} 

\titlerunning{Jia et al.} 
\authorrunning{Jia et al.} 
\author{Chengyou Jia, Minnan Luo, Caixia Yan,
Xiaojun Chang, Qinghua Zheng}
\institute{cp3jia@stu.xjtu.edu.cn}

\maketitle
\setlength{\belowcaptionskip}{-0.15cm}
\begin{abstract}
Recently, weakly supervised person search is proposed to discard human-annotated identities and train the  model with only bounding box annotations. A natural way to solve this problem is to separate it into detection and unsupervised re-identification (Re-ID) steps. However, in this way, two important clues in unconstrained scene images are ignored. On the one hand, existing unsupervised Re-ID models only leverage cropped images from scene images but ignore its rich context information. On the other hand, there are numerous unpaired persons in real-world scene images. Directly dealing with them as independent identities leads to the long-tail effect, while completely discarding them can result in serious information loss. In light of these challenges, we introduce a \textbf{C}ontext-\textbf{G}uided and \textbf{U}npaired-\textbf{A}ssisted (\textbf{CGUA}) weakly supervised person search framework. Specifically, we propose a novel \textbf{C}ontext-\textbf{G}uided \textbf{C}luster (\textbf{CGC}) algorithm to leverage context information in the clustering process and an \textbf{U}npaired-\textbf{A}ssisted \textbf{M}emory (\textbf{UAM}) unit to distinguish unpaired and paired persons by pushing them away. Extensive experiments demonstrate that the proposed approach can surpass the state-of-the-art weakly supervised methods by a large margin (more than 5\% mAP on CUHK-SYSU). Moreover, our method achieves comparable or better performance to the state-of-the-art supervised methods by leveraging more diverse unlabeled data. Codes and models will be released soon.
\end{abstract}

\section{Introduction}
Person search \cite{xiaoli2017joint,Zheng_2017_CVPR} aims to locate a query person in a gallery of unconstrained scene images, which can be viewed as a joint task of person detection and person re-identification (Re-ID). Without the requirement on given precise bounding boxes, person search is more suitable for real-world applications than the Re-ID task. In recent years, supervised person search methods \cite{Chen_2018_eccv,chang2018rcaa,han2019re,chen2020norm,wang2020tcts,li2021sequential,alignps} have achieved impressive performance, which rely on human-annotated bounding boxes and identities to train their models. However, collecting large-scale and wise-paired dataset with person identity annotations is prohibitively costly and labor-intensive. These difficulties lead researchers to explore new methods that can train the model with limited supervision.

\begin{figure}[t]
    \centering
    \includegraphics[width=0.8\textwidth]{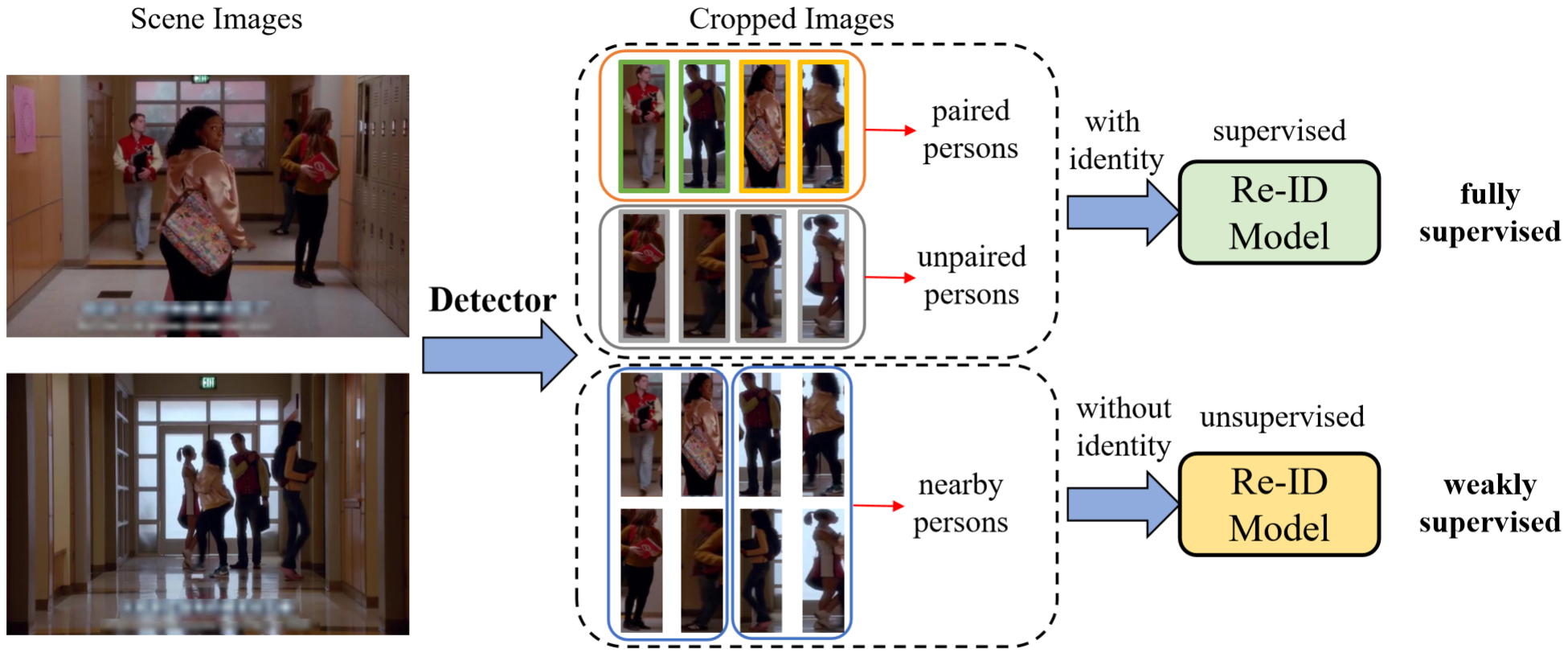}
    \caption{Illustration of fully-supervised setting and weakly-supervised setting. In the fully-supervised setting, both annotated bounding boxes and person identities are available. In the weakly-supervised setting, images only have bounding box annotations while person identities are lacking. Green and orange boxes index two different person identities, while gray boxes indicate unpaired person images.}
    \label{fig:intro}
\end{figure}

It is evident that the cost of annotating bounding boxes is much lower than annotating person identities. Therefore, a weakly supervised setting  of person search is proposed to train the model only with bounding box annotations \cite{yan2021exploring,han2021weakly}, thereby relieving the burden of human labeling. Intuitively, this task can be dealt with detection model and unsupervised Re-ID model independently, as shown in Fig. \ref{fig:intro}. Images cropped from scene images by the detector will be fed into an unsupervised Re-ID model to extract features for matching. However, directly combining the two models fails to exploit two kinds of information in scene images. First, the whole scene images include rich context information, \emph{e.g.}, nearby persons in Fig. \ref{fig:intro} or global scene, which has been proved to be useful for supervised person search \cite{dong2020bi,li2021sequential,yan2019learning}. However, existing unsupervised Re-ID methods only attend to cropped images while ignoring their context information. Second, as shown in Fig. \ref{fig:intro}, there are numerous unpaired persons in real-world scene images, which only appear once in the whole dataset. These unpaired persons also have been explored to further improve the performance \cite{xiaoli2017joint}, while existing unsupervised Re-ID methods fail to consider this.
\setlength{\belowcaptionskip}{-0.5cm}

In this paper, we propose a novel \textbf{C}ontext-\textbf{G}uided and \textbf{U}npaired-\textbf{A}ssisted weakly supervised person search framework, termed as \textbf{CGUA}. Our method consists of two main components: \textbf{C}ontext-\textbf{G}uided \textbf{C}luster (\textbf{CGC}) algorithm and \textbf{U}npaired-\textbf{A}ssisted \textbf{M}emory (\textbf{UAM}) unit. These two components are designed to leverage context information and unpaired persons respectively. 
Specifically, the proposed CGC algorithm is context-guided, which adopts the hybrid similarity (visual and context) to make the best of rich context information in scene images. Besides, based on the context property, CGC algorithm adds extra constraints to the clustering process to filter clustering results.
Another component, \emph{i.e.}, the UAM unit, is designed to take advantage of numerous unpaired persons in real-world scene images. Unlike previous methods that discard unpaired persons, the proposed UAM unit consists of a paired memory bank and an unpaired memory bank, which store features from paired and unpaired persons respectively. The unpaired memory bank assists the Re-ID model to learn more discriminative features by pushing these unpaired features away from paired features. Extensive experiments confirm that the proposed components yield a significant performance gain, with more than 15\% improvement of mAP and top-1 on the CUHK-SYSU dataset.
Our main contributions are summarized as follows:
\vspace{-0.25cm}
\begin{itemize}
    \item We develop a novel Context-Guided and Unpaired-Assisted weakly supervised person search framework. The proposed CGUA is trained without human-annotated identities to relieve the burden of human labeling, and thus is more suitable for real-world applications than supervised methods.
    \item We propose a context-guided cluster algorithm and an unpaired-assisted memory unit to leverage the rich context information and numerous unpaired persons respectively. The former makes the clustering more effective, while the latter reduces the confusion between paired and unpaired persons.
    \item The proposed method achieves the top-1 of 92.0\% and 86.9\% on CUHK-SYSU and PRW dataset respectively, which surpasses state-of-the-art weakly supervised methods by a large margin. Moreover, our method achieves comparable or better performance to the state-of-the-art supervised methods by leveraging more diverse unlabeled data.
\end{itemize}

\section{Related Works}{\label{sec:Related works}}

In this section, we briefly review the related works on the fields relevant to our study: person search and unsupervised person re-identification.

\subsection{Person Search}
Existing solutions to person search can be categorized into two groups: two-step methods and one-step methods. Two-step methods \cite{Zheng_2017_CVPR} solve the pedestrian detection task and person Re-ID task through two separated models. These methods focus on how to learn more discriminative Re-ID features based on the detection results. Methods of Mask-Guided \cite{Chen_2018_eccv}, Re-ID Driven \cite{han2019re}, and Task-Consistent \cite{wang2020tcts} are proposed to achieve this goal. In general, two-step methods obtain high performance but low efficiency in evaluation because they employ independent detection and Re-ID model. 
In contrast, one-step person search methods aim to solve the two tasks in a unified model, yielding higher efficiency than two-step methods. Xiao \emph{et al.} \cite{xiaoli2017joint} proposed the first one-step framework for person search, which is demonstrated to learn Re-ID features more effectively and efficiently. Beyond that, how to leverage context information \cite{chang2018rcaa,dong2020bi,yan2019learning}, relieve the conflict of shared features \cite{chen2020norm,li2021sequential}, and align features from multi-level \cite{lan2018person,alignps} are explored to achieve better performance. 

Although these supervised methods have achieved impressive progress, they usually rely on large-scale training data with the annotations of person identities. In real-world scenarios, collecting wise-paired data is difficult and annotating identities is labor-intensive. Thus, a weakly supervised setting \cite{yan2021exploring,han2021weakly} is proposed to train a person search model only with bounding boxes. The representative work, \emph{i.e.}, R-SiamNet \cite{han2021weakly}, introduces an effective weakly supervised person search model based on Siamese Networks. Despite its empirical success, this method ignores the context information in whole scene images and it fails to leverage diverse unlabeled data by adopting a one-step framework. 

\subsection{Unsupervised Person Re-ID}
Unsupervised person Re-ID aims to learn discriminative Re-ID features from unlabeled cropped images. Recent works are dominated by pseudo-label-based methods, which generate pseudo labels by a Re-ID feature clustering algorithm, such as $k$-means and DB-SCAN \cite{ester1996density}. HCT \cite{zeng2020hierarchical} proposed a hierarchical clustering-guided Re-ID method, which employs hierarchical clustering to generate pseudo-labels and conducts the training with these pseudo-labels. SPCL \cite{ge2020selfpaced} adopted a self-paced contrastive learning strategy to create more reliable clusters gradually. HHCL \cite{hu2021hard} proposed a hard-sample guided hybrid contrast learning framework to exploit the information of hard samples. Although these unsupervised person Re-ID methods achieve high performance, they are proposed for person Re-ID task and thus cluster features from cropped images. Applying them to person search will ignore the clue that each cropped image is from a known scene image. Whole scene images include rich context information which often plays an important role in improving Re-ID performance \cite{chang2018rcaa,dong2020bi,li2021sequential,yan2019learning}. Thus, we propose the CGC algorithm to explore context information in weakly supervised person search.

\section{Methodology}{\label{sec:Method}}

\textbf{Problem Definition and Overview.} In the setting of weakly supervised person search, we are given a set of $M$ scene images $S = \{S_1,...,S_M\}$ with their $N$ annotated bounding boxes $B = \{B_1,...,B_N\}$. From $S$ and $B$, we can obtain the cropped images set $I = \{I_1,...,I_N\}$. We use $Image(i)$ to indicate which scene image is the $i_{th}$ cropped image from. $V(i) = \{V_i^1, ..., V_i^{num(i)}\}$ refers to the set of cropped images in the $i_{th}$ scene image where $num(i)$ indicates the number of persons in the $i_{th}$ scene image. The goal of our task is to learn a detector $Detector(\cdot;\theta)$ from labeled scene images $S,B$ and a Re-ID encoder $
Encoder(\cdot;\theta)$ from unlabeled cropped images $I$. $\Gamma_i = Encoder(I_i;\theta)$ is the Re-ID feature of the cropped image $I_i$ generated by the Re-ID encoder.

\begin{figure*}[t]
    \centering
    \includegraphics[width=0.9\textwidth]{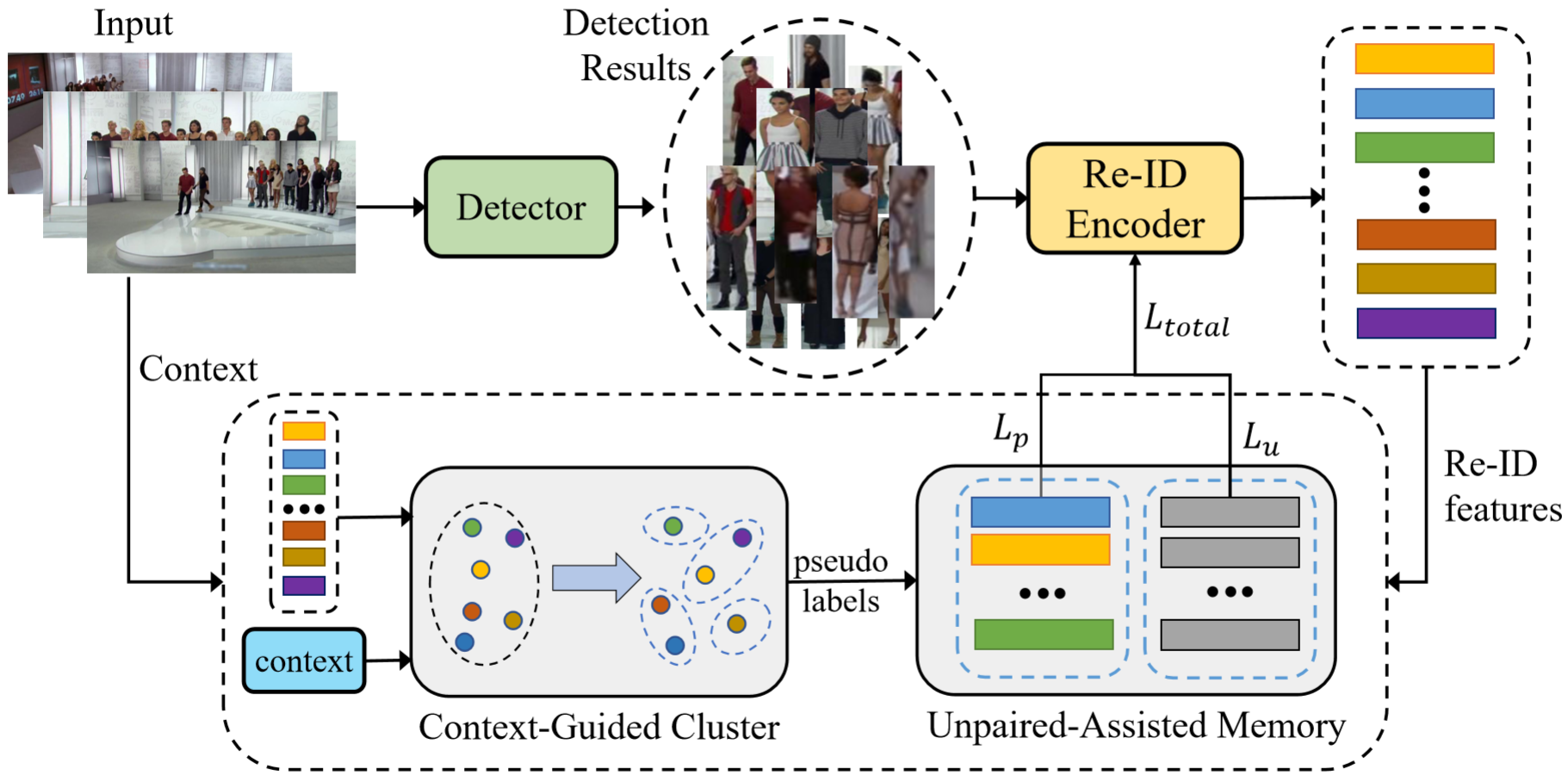}
    \caption{The overall architecture of the proposed two-step weakly supervised person search framework. First, the detection network takes the whole scene images as input to produce detection results. Then, the cropped images from detection results are fed into the Re-ID encoder and generate corresponding Re-ID features. Next, the proposed CGC algorithm generates pseudo labels by clustering these features with context information. After clustering, paired clusters that have more than two instance features are sent to initialize the paired memory bank with corresponding pseudo labels. Simultaneously, unpaired clusters that contain only one instance are used to initialize the unpaired memory bank. Finally, losses between query features and two banks $L_p,L_u$ are calculated and then back propagate to update the Re-ID model.}
    \label{fig:overview}
\end{figure*}

The overview of the proposed two-step person search architecture is presented in Fig. \ref{fig:overview}. Our model consists of two main components: Context-Guided Cluster (CGC) algorithm and Unpaired-Assisted Memory (UAM) unit. In the following, we provide the details of these two parts.
\setlength{\belowcaptionskip}{-0.55cm}
\setlength{\belowcaptionskip}{-0.15cm}
\vspace{-0.2cm}
\subsection{Context-Guided Cluster Algorithm}
Although similarity-based clustering algorithms \cite{ester1996density}
have been applied in the unsupervised person Re-ID task successfully \cite{zeng2020hierarchical,ge2020selfpaced,wang2020unsupervised}, they deal with cropped images individually and ignores the rich context information in scene images.
Besides, these clustering algorithms are parameter-based, \emph{e.g.}, the number of clusters or distance threshold. 
The choice of parameters is subjective and has to change for different datasets. 
In sight of these problems, we propose a novel Context-Guided Cluster (CGC) algorithm to leverage both visual and context information effectively. 
Notably, the proposed CGC algorithm is based on the efficient FINCH \cite{finch} clustering algorithm, which can handle large data efficiently without setting any hyper-parameters. 

\begin{figure}[t]
    \centering
    \includegraphics[width=0.8\textwidth]{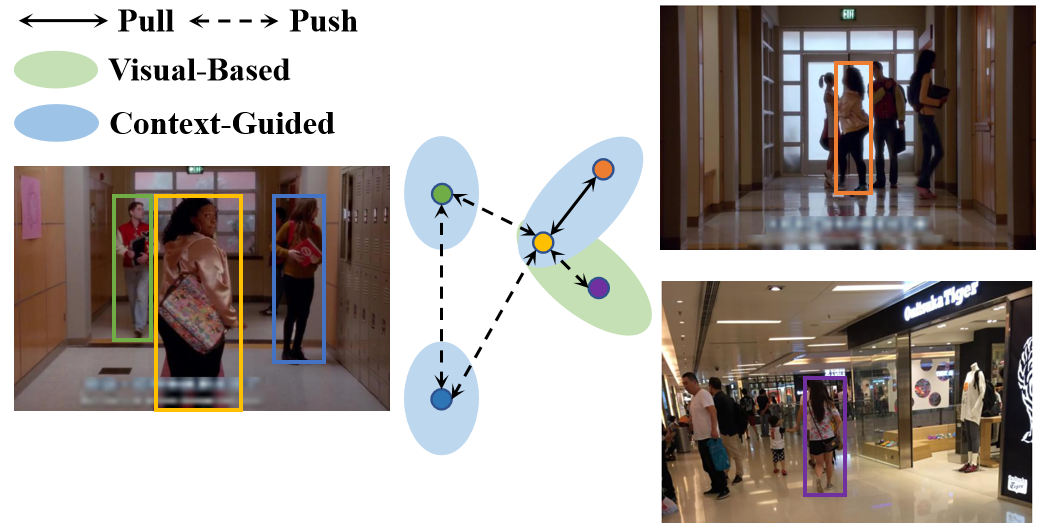}
    \caption{The illustration of two kinds of context information used in CGC algorithm.}
    \label{fig:cluster}
\end{figure}

\noindent \textbf{Context Information.} In the person search task, we have the natural weak label that each cropped image is from a known scene image. Based on that, we consider context information in two ways: Intra-image and Inter-image. ``Intra-image'' context information exists in the uniqueness property of person search: \textbf{\textit{persons from the same scene image cannot belong to the same cluster}}. This property adds extra constraints to the CGC algorithm to filter clustering results. ``Inter-image'' context information is from the clue that \textbf{\textit{persons tend to walk alongside same persons}}. The rich context information in potential co-travelers helps the CGC algorithm be more accurate than just relying on visual similarity. We present a specific example in Fig. \ref{fig:cluster}. The left part shows ``Intra-image'' context information: each person from the same scene image should be clustered into an independent category. The right part shows ``Inter-image'' context information: the person in the orange box has richer context similarity with the person in the yellow box than the person in the purple box. The green cycle represents clustering results depending on visual similarity while blue cycles denote results depending on context similarity. This example shows the context-guided strategy is more effective for clustering in person search.
\setlength{\belowcaptionskip}{-0.45cm}
\setlength{\belowcaptionskip}{-0.15cm}

\noindent \textbf{Clustering Algorithm.} Our CGC algorithm takes the clustering as a graph partitioning problem, where each node in the graph represents a cropped image and each partition in the graph means one cluster. We use a symmetric sparse matrix $A \in \mathbb{R}^{N \times N}$ to symbolize the graph where $A(i,j) \in \{0,1\}$ indicates whether there is a link between node $i$ and $j$. We define the matrix $A$ as follows: 
\begin{align}
\label{equ:consistency}
\begin{split}
A(i,j)=\left\{
\begin{array}{ll}
1 \quad & \text{if} \;  ( \; j = \kappa_{i}^{1} \; \text{or} \; \kappa_{j}^{1} = i \; \text{or} \;  \kappa_{i}^{1} = \kappa_{j}^{1} \; ) \\
& \text{and} \; Image(i) \neq Image(j) \\
0 \quad & \text{otherwise}
\end{array},
\right.
\end{split}
\end{align}
where $\kappa_{i}^{1}$ denotes the nearest neighbor of sample $i$. To better find $\kappa_{i}^{1}$, the CGC algorithm adopts not only visual similarity but also context similarity to compute the distances of all pairs. The visual similarity matrix is computed as
\begin{equation}
\label{equ:CosSim}
\centering
Q(i,j)  = CosSim(\Gamma_i, \Gamma_j),
\end{equation}
where $Q \in{\mathbb{R}}^{N \times N}$ denotes the cosine similarity matrix of $N$ features. Besides, we take maximum matching value as context similarity:
\begin{equation}
\label{equ:ContextSim}
\centering
K(i,j)  = \max_{m \in V(i),n \in V(j)} \; Q(m,n),
\end{equation}
where $K \in{\mathbb{R}}^{M \times M}$ denotes the context similarity matrix and $K(i,j)$ refers to the context similarity between scene image $S_i$ and $S_j$. $V(i)$ contains all cropped images in scene image $S_i$. 
The final hybrid similarity matrix is computed based on both visual and context:
\begin{equation}
\label{equ:FinalSim}
\centering
Q^{\prime}(i,j) = Q(i,j) + \lambda_{sim} \cdot K(Image(i), Image(j)),
\end{equation}
where $\lambda_{sim}$ is the trade-off coefficient between two kinds of similarity. Based on $Q^{\prime}$, we calculate $\kappa_{i}^{1}$ by
\begin{equation}
\label{equ:argmax}
\centering
\kappa_{i}^{1} = argmax \; Q^{\prime}(i,:) \;.
\end{equation}

Eq. \eqref{equ:consistency} limits that cropped images from the same scene image can not be clustered into same clusters. However, these cropped images may still be clustered through multi-hop links. Therefore, inspired by \cite{yan2021exploring},  we further filter these existing error clusters by only retaining the nearest instance to the cluster center.
The CGC algorithm is summarized in Algorithm \ref{alg:CGC}.
\begin{algorithm}[t] 
	\caption{Context-Guided Clustering Algorithm} 
	\label{alg:CGC} 
	\begin{algorithmic}[1] 
		\REQUIRE 
		Feature matrix $ \Gamma \in{\mathbb{R}}^{N \times d}$, where $N$ is total number of cropped images and $d$ is the feature dimension. 
		\STATE \textbf{Initialize:} Visual similarities matrix $Q \in{\mathbb{R}}^{N \times N}$, context similarities matrix $K \in{\mathbb{R}}^{M \times M}$, hybrid similarity matrix $Q^{\prime} \in{\mathbb{R}}^{N \times N}$, all values are set to 0.
		\STATE Compute visual similarities matrix $Q$ using Eq. \eqref{equ:CosSim};
		\STATE Compute context similarities matrix $K$ using Eq. \eqref{equ:ContextSim};
		\STATE Compute hybrid similarities matrix $Q^{\prime}$ using Eq. \eqref{equ:FinalSim};
		\STATE Compute first neighbors integer vector $\kappa^{1} \in{\mathbb{R}}^{N \times 1}$ using Eq. \eqref{equ:argmax};
		\STATE Given $\kappa^{1}$ compute $A \in{\mathbb{R}}^{N \times N}$ using Eq. \eqref{equ:consistency};
		\STATE	Given $A \in{\mathbb{R}}^{N \times N}$  get clusters $C = \{C_1, C_2, \cdots,C_{N_c-1}, C_{N_c}\}$ via graph partition;
		\STATE Filter error clusters that have  cropped images from the same scene images;
		
		\ENSURE Clustering result $C = \{C_1, C_2, \cdots, C_{N_c-1}, C_{N_c}\}$
	\end{algorithmic}
\end{algorithm}

\subsection{Unpaired-Assisted Memory Unit}
As shown in Fig. \ref{fig:intro}, there are lots of unpaired persons in real-world scene images. Dealing with each unpaired person as an independent identity leads to the imbalanced long-tail effect, which prevents the Re-ID model from learning discriminative features. However, completely discarding these unpaired data would result in serious information loss. Therefore, we propose an Unpaired-Assisted Memory (UAM) unit to take advantage of those unpaired persons.

Our UAM unit includes a paired memory bank $\mathbf{M}_p \in {\mathbb{R}}^{N_p \times d}$ and an unpaired memory bank $\mathbf{M}_u \in {\mathbb{R}}^{N_u \times d}$ to store embeddings of all paired instance features and unpaired instance features respectively, where $N_p$ and $N_u$ denote the number of paired clusters and unpaired clusters with $d$ being the feature dimension. We adopt contrastive learning method to minimize the distance between same identities and maximize the distance between different identities. In UAM unit, we design the contrastive loss in two ways: paired cluster contrastive loss $L_{p}$ and unpaired cluster contrastive loss $L_{u}$. The former focuses on increasing the intra-identity compactness and inter-identity separability of paired instances while the latter aims to pull unpaired instances away from paired instances. The overall loss function of the UAM unit is formulated as:
\begin{equation}
\label{equ:ReidLoss}
\centering
L_{reid} = \lambda_{reid} \cdot L_{p} + (1-\lambda_{reid}) \cdot L_{u},
\end{equation}
where $\lambda_{reid}$ is a balancing factor. We describe details of $L_{p},L_{u}$ in the following.

\noindent \textbf{Paired Cluster Contrastive Loss $L_{p}$.}
As mentioned in \cite{dai2021cluster}, instance-level memory dictionary techniques \cite{wang2020unsupervised,ge2020selfpaced} suffer from inconsistency in the updating progress of each cluster. Thus, in this paper, we compute the loss and update the memory dictionary $\mathbf{M}_p$ both at the cluster level. For each query feature $q$, paired cluster contrastive loss is calculated as:

\begin{equation}
\label{equ:clusterLoss}
\centering
L_{cluster}=-\log\frac{\exp({q \cdot c^+} / {\tau_c})}{\sum_{i=1}^{N_p}\exp({q \cdot c^i} / {\tau_c})},
\end{equation}
where $ \{c^1, c^2,\cdots, c^{N_p}\} $ is the set of cluster centroids and $c^+$ is a positive cluster centroid for $q$; $\tau_c$ denotes the temperature hyper-parameter that controls the scale of similarities. $c^i$ in memory bank $M_p$ is updated as:
\begin{equation}
\label{equ:clusterUpdate}
\centering
c^i \leftarrow m c^i + (1-m) \bar{c}^i,
\end{equation}
where $\bar{c}^i$ is the average of $i_{th}$ identity instance features in a mini-batch and $m$ is the momentum updating factor.

In addition, the exploitation of hard-samples has been demonstrated \cite{hu2021hard} to be very effective in improving performance for unsupervised Re-ID model. We follow a hard-sample mining scheme and hard-based loss as \cite{hu2021hard} to help our model learn more discriminative features. The hard-based loss is computed as:
\begin{equation}
\label{equ:hardLoss}
\centering
L_{hard}=-\log\frac{\exp({q \cdot c_{hard}^+} / {\tau_c})}{\sum_{i=1}^{N_p}\exp({q \cdot c^i_{hard}} / {\tau_c})},
\end{equation}
where $c_{hard}^+$ is the hard positive instance feature which has the lowest
similarity with query $q$ in the same cluster; $c^i_{hard}$ is the hard negative instance feature that has the highest similarity with query $q$ but belongs to different clusters. Finally, the total paired cluster contrastive loss of the UAM unit is defined as:

\begin{equation}
\label{equ:pairedLoss}
\centering
L_{p}=L_{cluster} + L_{hard}.
\end{equation}

\noindent \textbf{Unpaired Cluster Contrastive Loss $L_{u}$.} Because there is only one instance in each unpaired cluster, we compute unpaired cluster contrastive loss and update the memory dictionary $\mathbf{M}_u$ both at the instance level. For each query feature $q$, the unpaired cluster contrastive loss is calculated as follows:
\begin{equation}
\label{equ:unclusterLoss}
\centering
L_{u}=-\log\frac{\exp({q \cdot u^*} / {\tau_c})}{\sum_{i=1}^{N_u}\exp({q \cdot u^i} / {\tau_c})},
\end{equation}
where $ \{u^1, u^2,\cdots, u^{N_u}\} $ are unpaired instance features. Each feature indexes an independent cluster. It's worth noting that all unpaired features will be only used for updating $M_u$ and will not be used as a query feature $q$. Each query feature is from a paired cluster and thus there are no positive instance features in memory bank $M_u$ with feature $q$. So we randomly choose a feature $u^*$ from $M_u$ as the positive instance. This random choice strategy is demonstrated to be very effective, which means unpaired memory bank only assists paired cluster features to computed loss but not be a subject. $u^i$ in unpaired memory bank $M_u$ is updated as follows:

\begin{equation}
\label{equ:unclusterUpdate}
\centering
u^i \leftarrow m u^i + (1-m) u^i_{new},
\end{equation}
where $u^i_{new}$ is the new instance feature generated by the updated Re-ID model for unpaired cluster $i$.

\subsection{Training and Inference Details}
\textbf{Training.} We adopt a two-step training mechanism to optimize the network parameters of detector $Detector(\cdot;\theta)$ and Re-ID model $Encoder(\cdot;\theta)$ respectively. We employ the classical Faster R-CNN \cite{rennips15fasterrcnn} as our detector and follow its training strategy. The loss for the detector is formulated as:
\begin{equation}
\label{equ:detectorLoss}
\centering
L_{detector} = L_{reg} + L_{cls},
\end{equation}
where $L_{reg}$ and $L_{cls}$ denote the same regression loss and classification loss as in Faster R-CNN respectively.
For the Re-ID model $Encoder(\cdot;\theta)$, we use unlabeled images set $I$ and their pseudo labels to train it with the contrastive loss $L_{reid}$.

\noindent\textbf{Inference.} Given a cropped query image $I_q$ and a gallery set of scene images $S^g$, we firstly use $Detector(\cdot;\theta)$ to generate bounding box predictions $B^*$ for scene images in $S^g$. Subsequently, all person images cropped from predictions $B^*$ are fed into $Encoder(\cdot;\theta)$ to extract Re-ID features. All gallery features make up the set $\Gamma^*$ and $\Gamma_q$ being the feature of query $I_q$. Finally, we search the set $\Gamma^*$ to retrieve the most similar matches to $\Gamma_q$ based on their cosine similarity. The most similar feature is selected as the final result of person search.

\section{Experiments}{\label{sec:Experiments}}

\subsection{Experimental Setup}

\noindent \textbf{Datasets.} We evaluate the proposed model on two benchmark datasets for person search, \emph{i.e}, CUHK-SYSU \cite{xiaoli2017joint} and PRW \cite{Zheng_2017_CVPR}. Specifically, CUHK-SYSU dataset consists of 18,184 images and 96,143 annotated pedestrian bounding boxes with 8,432 identities. We follow the standard train/test split: 11,206 images for training while 2,900 query images and 6,978 gallery images for testing. PRW dataset contains 11,816 video frames and 43,110 annotated pedestrian bounding boxes with 932 identities. It is more challenging because each identity has more ground truth bounding boxes (36.8 \textit{vs} 2.8 in CUHK-SYSU). We also adopt the standard train/test split: 5,704 images for training while 2,057 query images and 6,112 gallery images for testing.

\noindent \textbf{Evaluation Protocols.} We evaluate the proposed method with two widely adopted protocols \cite{xiaoli2017joint}, \emph{i.e}, mean Average Precision (mAP) and Cumulative Matching Characteristics (CMC). For the mAP, Average Precision(AP) is computed for each query based on the precision-recall curve and results are averaged to calculate the mAP. For the CMC metric, a matching is counted if there is at least one of the top-$k$ predicted bounding boxes overlapping with the ground truth. A correct overlap means the IOU (Intersection Over Union) with ground truth is larger than or equal to 0.5.

\noindent \textbf{Implementation Details.} The detector of our model is built upon Faster R-CNN \cite{rennips15fasterrcnn}. Its backbone is ResNet-50\cite{he2016deep} with FPN\cite{lin2017feature}. We adopt a multi-scale training strategy, where the longer side of the input image is fixed to 1333 pixels and the shorter side is resized from 640 to 800 pixels randomly. The detector is trained for 24 epochs with the learning rate multiplied by 0.1 at 16 and 22 epochs. Finally, it achieves the performance with the mAP of 93.3\% and 94.1\% on the CUHK-SYSU and PRW, respectively.

For the Re-ID model, we adopt ResNet-50 \cite{he2016deep} as the backbone and initialize the model with parameters pre-trained on ImageNet \cite{5206848}. All sub-module layers after layer-4 are removed and generalized mean pooling (GeM)\cite{Radenovi2019FineTuningCI} is added. Finally, a L2-normalization layer
is adopted to produce 2048-dimensional features which are treated as Re-ID features. In the training stage, each cropped input image is resized to 256 × 128. 
The batch size is set to 64 for both the two datasets, while the total iterations in each epoch are 5000 for CUHK-SYSU and 500 for PRW. 
The Adam optimizer is adopted with an initial learning rate 3.5e-4 that is reduced to 1/10 every 10 epochs. We submit the evaluation of parameters $\lambda_{sim},\lambda_{reid}$ and implemented codes in supplementary material.
\setlength{\belowcaptionskip}{-0.15cm}
\begin{table}[t]
  \centering
    \caption{Comparison of mAP and top-1 accuracy with state-of-the-art methods.}
  \begin{tabular}{l|l|c|cc|cc}
    \toprule
    \multicolumn{2}{c|}{\multirow{2}{*}{Methods}} & \multirow{2}{*}{\quad Reference \quad} & \multicolumn{2}{|c|}{\ \  CUHK-SYSU  \quad} & \multicolumn{2}{c}{ PRW } \\
    \multicolumn{2}{c|}{}  & & \quad mAP & top-1 \quad & \quad mAP \quad & top-1 \quad \\
    \midrule
    \multirow{14}{*}{\rotatebox{90}{supervised}} 
                            & OIM   \cite{xiaoli2017joint} & CVPR'17& 75.5 & 78.7 & 21.3 & 49.4 \\
                            & RCAA  \cite{chang2018rcaa} & ECCV'18   & 79.3 & 81.3 &   -  &  -   \\
                            & MGTS  \cite{Chen_2018_eccv}   & ECCV'18& 83.0 & 83.7 & 32.6 & 72.1 \\
                            & CLSA  \cite{lan2018person} & ECCV'18 & 87.2 & 88.5 & 38.7 & 65.0 \\
                            & CTXGraph  \cite{yan2019learning} & CVPR'19 & 84.1 & 86.5 & 33.4 & 73.6 \\
                            & HOIM  \cite{chen2020hoim} & AAAI'20     & 89.7 & 90.8 & 39.8 & 80.4 \\
                            & BINet  \cite{dong2020bi} & CVPR'20& 90.0 & 90.7 & 45.3 & 81.7 \\
                            & NAE  \cite{chen2020norm} & CVPR'20   & 91.5 & 92.4 & 43.3 & 80.9 \\
                            & RDLR  \cite{han2019re}  &
                            ICCV'19 & 93.0 & 94.2 & 42.9 & 70.2 \\
                           & AlignPS \cite{alignps} & CVPR'21 & 93.1 & 93.4 & 45.9 & 81.9 \\
                           & SeqNet  \cite{li2021sequential}&
                           AAAI'21& 93.8 & 94.6 & 46.7 & 83.4 \\
                           & DKD \cite{zhang2021diverse} &AAAI'21&93.1&94.2&50.5&87.1\\
                           & AGWF \cite{Han_2021_ICCV} &
                          ICCV'21 &93.3&94.2&\textbf{53.3}&\textbf{87.7}\\
                           & TCTS  \cite{wang2020tcts} &
                           CVPR'20 & \textbf{93.9} & \textbf{95.1} & 46.8 & 87.5 \\
    \midrule
    \multirow{4}{*}{\rotatebox{90}{weakly-su}} &  Context-Aware \cite{han2021context}& ArXiv'21   & 81.1 & 83.2 & 41.7 & 86.0 \\
                            & CGPS \cite{yan2021exploring} &
                            AAAI'22& 80.0 & 82.3 & 16.2 & 68.0 \\
                            & R-SiamNet \cite{han2021weakly} & ICCV'21    & 86.0 & 87.1 & 21.2 & 73.4 \\
 
                            & \textbf{CGUA(Ours)} & This paper    & \textbf{91.0}  & \textbf{92.2}  & \textbf{42.7} & \textbf{86.9} \\
    \bottomrule
  \end{tabular}
  \label{tab:SOTA}
\vspace{-0.4cm}
\end{table}

\begin{figure}[t]
	\centering 	
	\subfigure[CUHK-SYSU]{\label{components_a}		
		\includegraphics[width=0.45\linewidth]{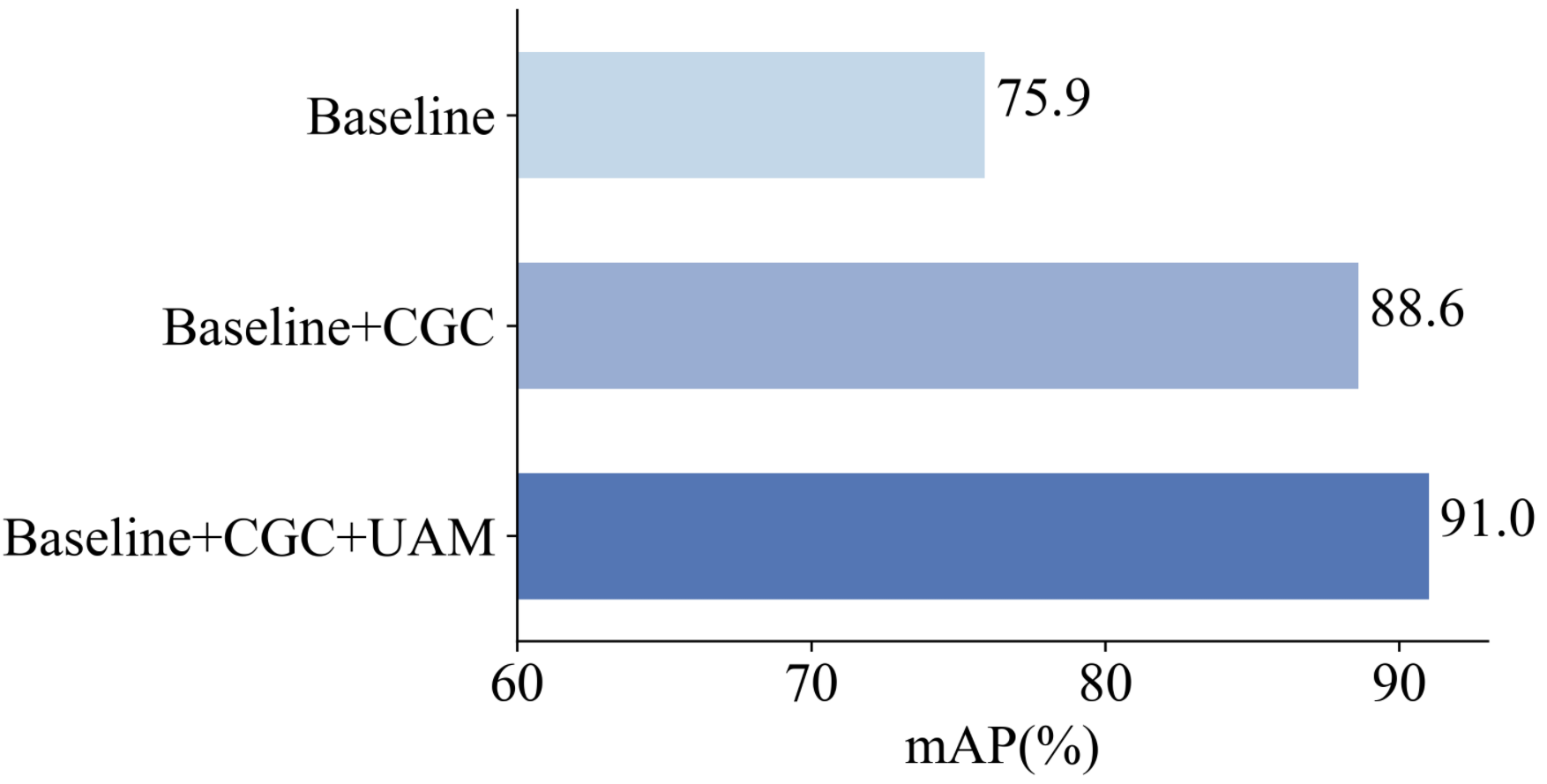}}\hspace{10pt}
	\subfigure[PRW]{\label{components_b}			
		\includegraphics[width=0.45\linewidth]{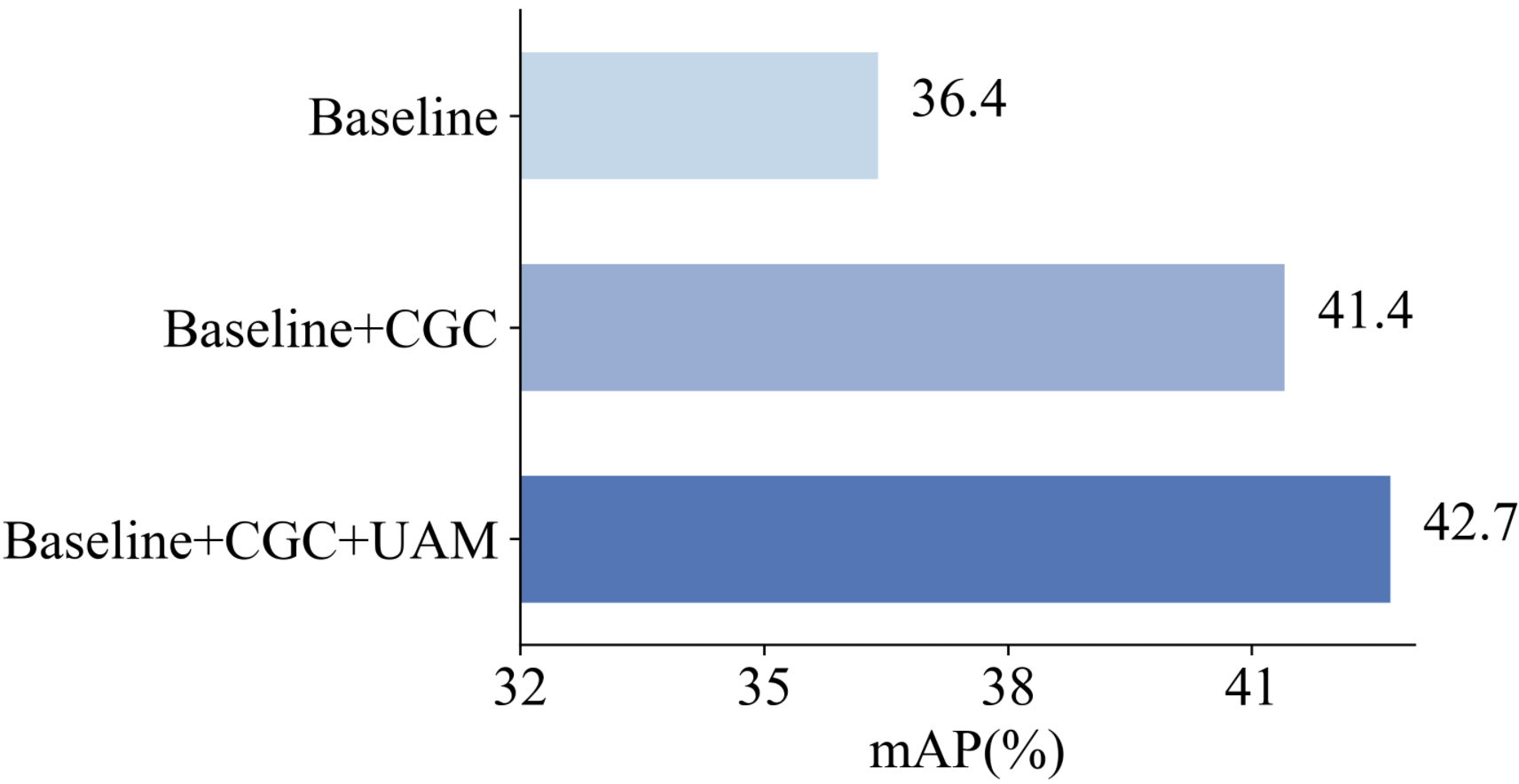}}
	\caption{Comparative results on CUHK-SYSU and PRW with different proposed  components, \emph{i.e.}, Context-Guided Cluster (CGC), Unpaired-Assisted Memory (UAM).}	
	\label{fig:CGC+UAM}	
	\vspace{0.1cm}
\end{figure}

\begin{table}[t]
 	\renewcommand{\arraystretch}{1.2}
 	\tabcolsep=5 pt	
 	\begin{center}
 		\caption{Comparison to different designs of CGC algorithm.}\label{tab:CGC}
 		\begin{tabular}{lcccccc}
 			\toprule
 			\multirow{2}*{Cluster}&\multirow{2}*{inter}&\multirow{2}*{intra}&\multicolumn{2}{c}{CUHK}&\multicolumn{2}{c}{PRW}\\
 			\cmidrule{4-7}
 			&&&mAP&rank1&mAP&rank1\\
 			\midrule	
 			DB-SCAN \cite{ester1996density} &&&59.6&59.3&40.7&86.5\\
 			FINCH \cite{finch}&&&75.9&77.0&36.4&84.4\\
 			FINCH w/inter&$\surd$&&76.7&78.3&37.5&87.1\\
 			FINCH w/intra&&$\surd$&87.3&88.0&39.4&85.8\\
 			\textbf{CGC}&$\surd$&$\surd$& \textbf{88.6}  & \textbf{89.8} & \textbf{41.4}  & \textbf{88.1}  \\
 			\bottomrule
 		\end{tabular}	
 	\end{center}
\vspace{-0.6cm}
 \end{table} 

\subsection{Comparison with The State-of-the-art}

In this section, we present the mAP and top-1 performance on two benchmarks in Table \ref{tab:SOTA} to compare the proposed framework with current state-of-the-art methods on person search. The results of supervised methods are shown in the upper block while weakly-supervised methods are presented in the lower block.

\begin{figure}[t]
	\centering 	
	\subfigure[Weakly supervised methods]{\label{Visualization_a}			
		\includegraphics[width=0.42\linewidth]{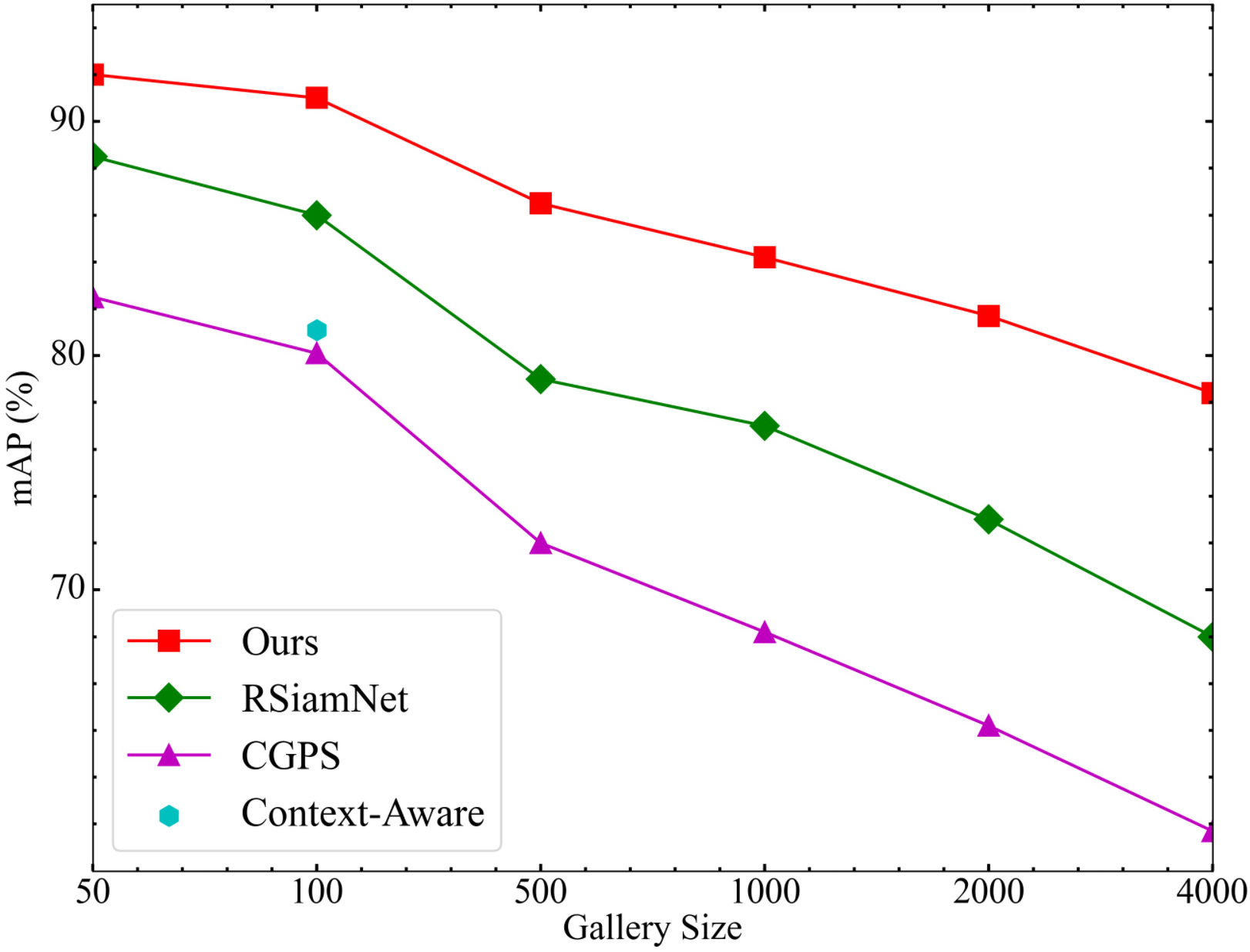}}
	\subfigure[Supervised methods]{\label{Visualization_b}		
		\includegraphics[width=0.42\linewidth]{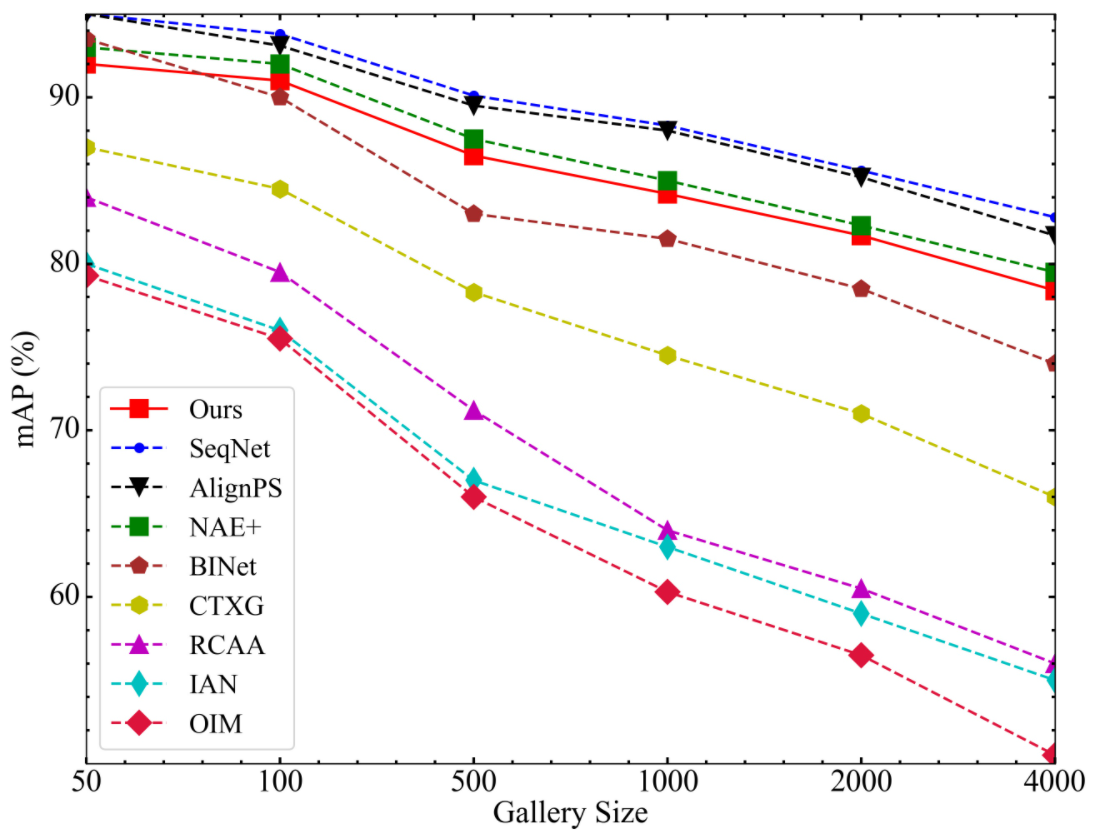}}\hspace{10pt}
	\vspace{-0.3cm}
	\caption{Comparison to different methods under varying gallery sizes.}	
	\label{fig:gallerySize}	
\vspace{-0.2cm}
\end{figure}

\noindent \textbf{Evaluation on CUHK-SYSU.}
As shown in Table \ref{tab:SOTA}, the proposed method achieves 91.0\% on
mAP and 92.2\% on top-1, which outperforms the state-of-the-art weakly-supervised methods by a large margin (more than 5\%). It is noteworthy that our method outperforms most fully-supervised methods, \emph{e.g.}, OIM \cite{xiaoli2017joint}, HOIM \cite{chen2020hoim} and BINet \cite{dong2020bi}, although these methods are trained with extra identity annotations. Besides, we further present the performance on CUHK-SYSU under varying gallery sizes of $[50,100,500,1000,2000,4000]$ in Fig. \ref{fig:gallerySize}. A larger gallery size corresponds to larger search scope, meaning that more distracting people are involved in matching, which makes person search more difficult. When the gallery size
increases, our method still outperforms all existing weakly-supervised methods by notable margins, which indicates our method can handle more challenging situations and is more suitable for real-world applications.

\noindent \textbf{Evaluation on PRW.} PRW is more challenging because of less training data and larger gallery size. Although our method achieves the best performance among weakly-supervised methods, the mAP of 42.7\% is still unsatisfactory. We infer that this is caused by underfitting and thus pre-train the model on more unlabeled data. Results in Table \ref{tab:moredata} show the mAP is improved by a large margin (more than 10\%) on PRW, which reveals the potential of our model.

\subsection{Ablation Study}

\noindent \textbf{Effectiveness of Different Components.} 
We conduct extensive quantitative analysis for the key components, \emph{i.e.}, CGC and UAM, in the proposed model by leaving one component out of our framework. It should be pointed out that the UAM unit can not be applied to the baseline because the original clustering algorithm can not produce clusters with only one instance. As shown in Fig. \ref{fig:CGC+UAM}, the baseline method only obtains 75.9\% and 36.4\% mAP on two benchmarks. The proposed CGC algorithm improves the baseline by 12.7\% and 5.0\% on mAP. Furthermore, the final model yields 15.1\% and 6.3\% improvements by combining both the CGC algorithm and UAM unit.
 

\begin{figure}[t]
	\centering 	
	\subfigure[FINCH algorithm]{\label{5b}			
		\includegraphics[width=0.42\linewidth]{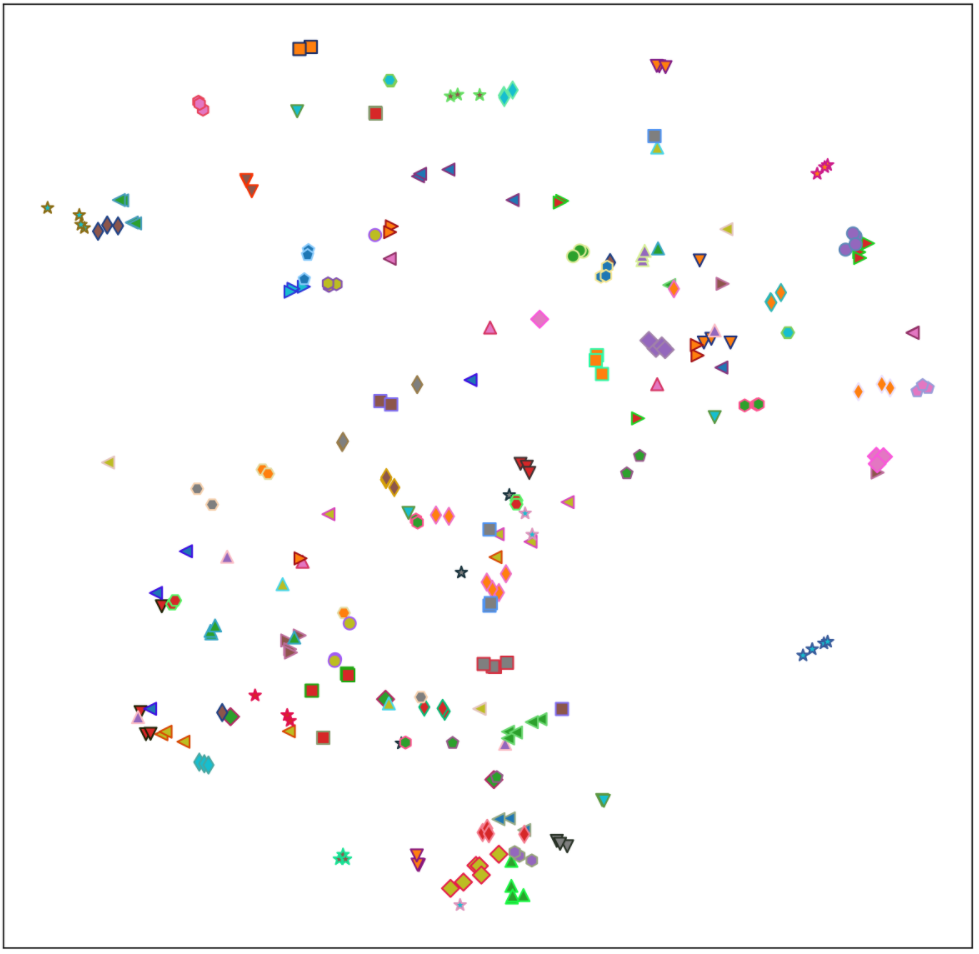}}
	\subfigure[CGC algorithm]{\label{5a}		
		\includegraphics[width=0.42\linewidth]{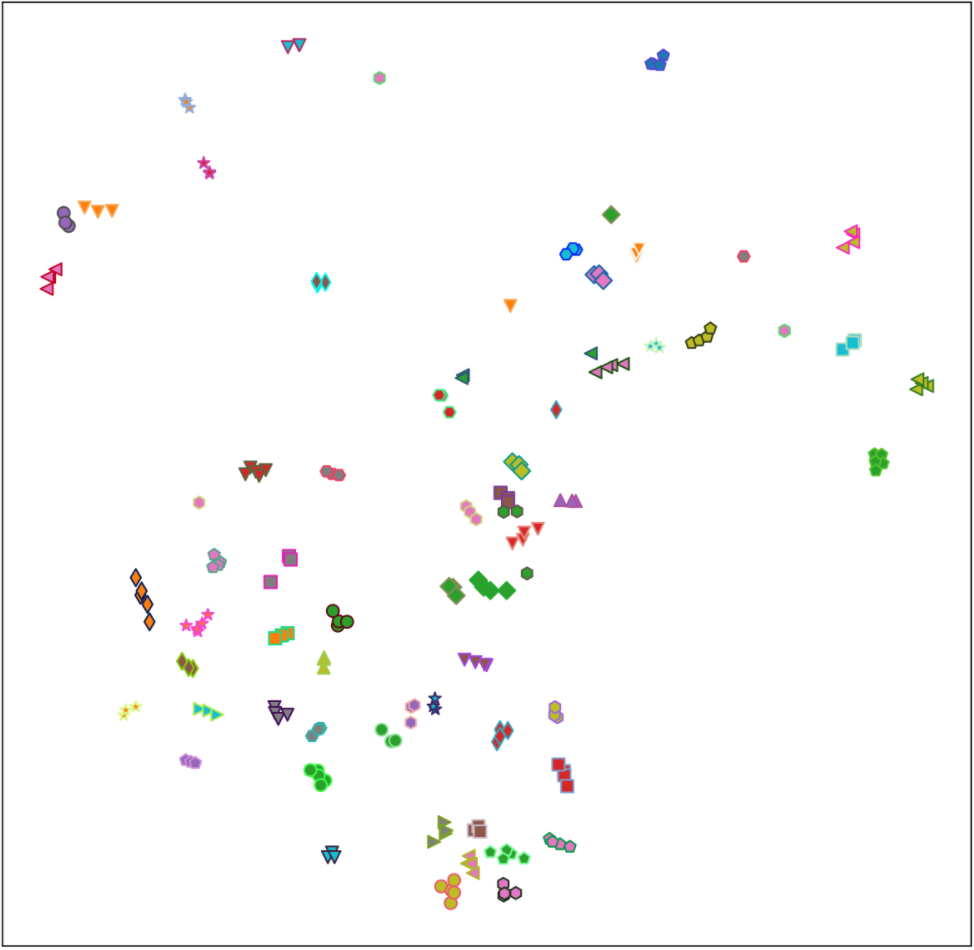}}
	\vspace{-0.3cm}
	\caption{Visualization of Re-ID features and clustering results by FINCH and CGC algorithm  respectively. We only display clusters that have more than 2 instances.}	
	\label{fig:tsne}	
\vspace{-0.5cm}
\end{figure}

\begin{figure}[t]
  \centering
   \includegraphics[width=0.9\linewidth]{{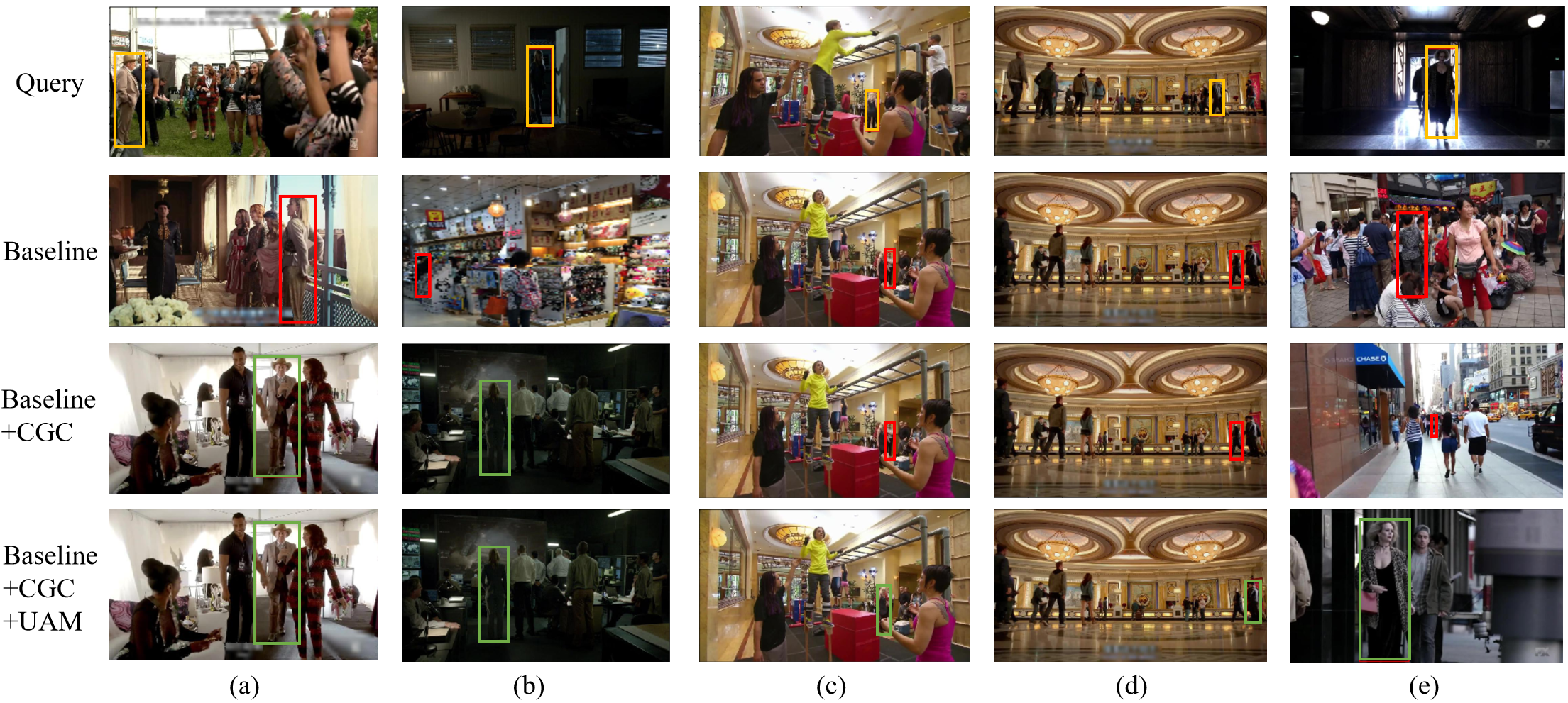}}
    \vspace{-0.4cm}
   \caption{Top-1 search results for several representative cases. The bounding boxes in yellow denote the queries while green and red denote the correct and wrong results.}
   \label{fig:Case}
\vspace{-0.0cm}
\end{figure}

\noindent \textbf{Design of the CGC algorithm.} 
In this part, we explore the design of the proposed CGC algorithm. We provide five different clustering variants: DB-SCAN in \cite{ester1996density,hu2021hard}, FINCH in \cite{finch}, FINCH with ``inter-image'' context information, FINCH with ``intra-image'' context information and CGC that considers both of two kinds of context information. As shown in Table \ref{tab:CGC}, although DB-SCAN achieves slightly better performance on PRW, it has a terrible performance on CUHK-SYSU. The results indicate that FINCH is more suitable for different datasets since it avoids the dependency of parameters. Further, both ``inter-image'' and ``intra-image'' context information are verified to be essential. Finally, CGC algorithm boosts the performance by more than 12\% and 5\% mAP on CUHK-SYSU and PRW respectively, compared to the original FINCH algorithm.


\begin{table}[t]
 	\renewcommand{\arraystretch}{1.2}
 	\tabcolsep=5 pt	
 	\begin{center}
 		\caption{Utilization of diverse unlabeled data.}\label{tab:moredata}
 		\begin{tabular}{lcccccc}
 			\toprule
 			\multirow{2}*{Extra}&\multirow{2}*{Box}&\multirow{2}*{Identity}&\multicolumn{2}{c}{CUHK}&\multicolumn{2}{c}{PRW}\\
 			\cmidrule{4-7}
 			Dataset&&&mAP&top-1&mAP&top-1\\
 			\midrule	
 			- &-&-&91.0&92.2&42.7&86.9\\
 			
 			INRIA \cite{1467360} &$\times$&$\times$&91.3&92.6&44.0&87.0\\ 			INRIA \cite{1467360}&$\surd$&$\times$&91.5&92.8&44.5&87.2\\
 			Market \cite{zheng2015scalable}&-&$\times$&\textbf{93.1}&\textbf{93.7}&\textbf{54.5}&\textbf{89.5}\\
 			\bottomrule
 		\end{tabular}	
 	\end{center}
\vspace{-0.8cm}
\end{table} 
\begin{table}[t]
  \centering
    \caption{Runtime comparison of different models.\label{tab:runtime}}
  \begin{tabular}{l|l|c|c|c}
    \toprule
    \multicolumn{2}{c}{\multirow{1}{*}{Methods}} &  \multicolumn{1}{|c|}{\quad \ Supervised \quad \ } &\multicolumn{1}{|c|}{\quad \ GPU \quad \ } & \multicolumn{1}{c}{\quad Time(ms)} \\
    \midrule
    \multirow{6}{*}{\rotatebox{90}{one-step}} 
                            & OIM   \cite{xiaoli2017joint} & $\surd$ & V100 & 118  \\
                            & NAE  \cite{chen2020norm}      & $\surd$ & V100 & 83  \\
                           & NAE+  \cite{chen2020norm}      & $\surd$ & V100 & 98  \\
                           & SeqNet  \cite{li2021sequential}& $\surd$ & V100 & 86  \\
                           & AlignPS  \cite{wang2020tcts}    & $\surd$& V100 & \textbf{61}  \\
                           & CGPS  \cite{yan2021exploring}    &$\times$ & V100 & 68 \\
    \midrule
    \multirow{4}{*}{\rotatebox{90}{two-step}}
                            & MGTS \cite{Chen_2018_eccv} & $\surd$ & K80 & 1269 \\
                            & FRCNN+SBL \cite{yan2021exploring} & $\times$ & V100 & 101 \\
                            & FRCNN+SPCL \cite{yan2021exploring} & $\times$ & V100 & 100 \\
                            & \textbf{Ours}     & $\times$  & V100  & \textbf{68}  \\
    \bottomrule
  \end{tabular}
\vspace{-0.5cm}
\end{table}

\subsection{Qualitative Analysis}

\noindent \textbf{Clustering Results.} To further demonstrate the
effectiveness of the proposed CGC algorithm, we utilize t-SNE \cite{van2008visualizing} to visualize the Re-ID features and clustering results. Specifically, we generate Re-ID features for 500 random samples using models based on FINCH and CGC algorithm. These features are clustered by corresponding algorithms and processed with t-SNE from 2048-dim to 2-dim for visualization purposes. Compared to FINCH, as shown in Fig. \ref{fig:tsne}, features generated by the CGC algorithm have better intra-class compactness and inter-class separability. These qualitative results further demonstrate that the proposed CGC algorithm can produce more discriminative Re-ID features.

\noindent \textbf{Person Search Results.} For qualitative analysis of person search results, we present some qualitative results of our method (\emph{i.e.}, baseline+CGC+UAM) and its two variants (\emph{i.e.}, baseline and baseline+CGC) in Fig. \ref{fig:Case}. These results can explain why the proposed components are effective. For example, case (a), (b) and (e) show that our method without CGC gets wrong top-1 results. These wrong results have large visual similarity with queries, which can easily mislead the model. In comparison, the final model can get correct results because it further leverages the context similarity. Case (c), (d) and (e) show the method without UAM is easy to confuse queries and unpaired persons while the final model distinguishes them well, which indicates the proposed UAM unit can indeed help the model discriminate paired and unpaired persons.


\vspace{-0.3cm}
\subsection{Advantage Analysis}
In supervised person search, one-step methods are superior to two-step methods because of their effectiveness and efficiency. However, in the weakly supervised setting, our method is better than one-step methods in some ways.

\noindent \textbf{Utilization of Diverse Unlabeled Data.} For one-step models, the input is a whole scene image, which prevents them from leveraging cropped images. Another disadvantage of one-step models is that their detection performance compromises with the Re-ID performance, which can't make use of totally unlabeled scene images (\emph{i.e.}, the images without box and identity annotations), since their person search models can't predict precise bounding boxes. In contrast, our two-step model can take advantage of diverse unlabeled data because of its two-stage input and high detection performance. Table \ref{tab:moredata} shows three kinds of unlabeled data can indeed boost the performance. Particularly, our method achieves comparable or better performance to the state-of-the-art supervised methods by leveraging cropped image dataset (\emph{i.e.}, Market \cite{zheng2015scalable}), which demonstrates the potential of our model to utilizing unlabeled data.

\noindent \textbf{Efficiency Comparison.} Table \ref{tab:runtime} shows that our two-step method is as efficient as one-step methods, no matter supervised or weakly supervised methods. Although our method employs two individual models, the robustness of our Re-ID model helps mitigate the impact of inaccurate detection results, which reduces the complexity of our detector and improves the inference speed of our model.

\vspace{-0.3cm}
\section{Conclusion}
In this paper, we proposed a novel context-guided and unpaired-assisted weakly supervised person search framework. Our method is able to train the model without human-annotated person identities and is demonstrated to take advantage of diverse unlabeled data. Benefiting from the proposed CGC algorithm and UAM unit, our method achieves the state-of-the-art performance on two benchmarks and the gap with supervised methods becomes narrowed. Our method provides a novel perspective for weakly person search research and we hope future work towards better performance with less labeled data.

\clearpage
%
%
\bibliographystyle{splncs04}
\bibliography{egbib}

\clearpage
\section{Appendix}
The Appendix accompanies our paper ``CGUA: Context-Guided and Unpaired-Assisted Weakly Supervised Person Search'', including more experiment results and implementation details. We also submit source codes as supplementary material to reproduce.

\subsection{Experiments}
We conducted experiments on the CUHK-SYSU dataset to explore the influence of different parameters $\lambda_{sim},\lambda_{reid}$  on our method.

\begin{figure}
	\centering 	
	\subfigure[Evaluation of $\lambda_{sim}$]{\label{components_a}		
		\includegraphics[width=0.45\linewidth]{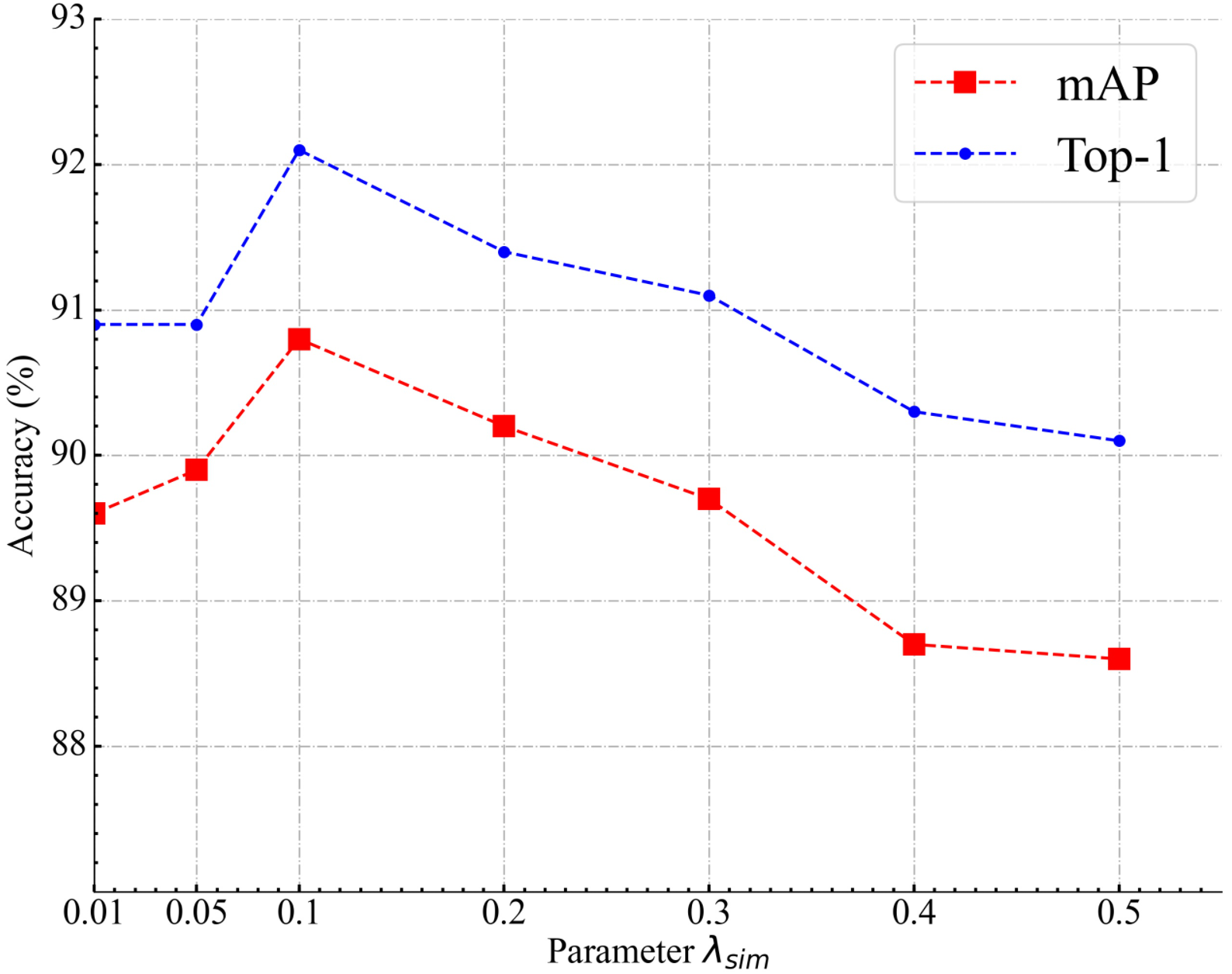}}\hspace{10pt}
	\subfigure[Evaluation of $\lambda_{reid}$]{\label{components_b}			
		\includegraphics[width=0.45\linewidth]{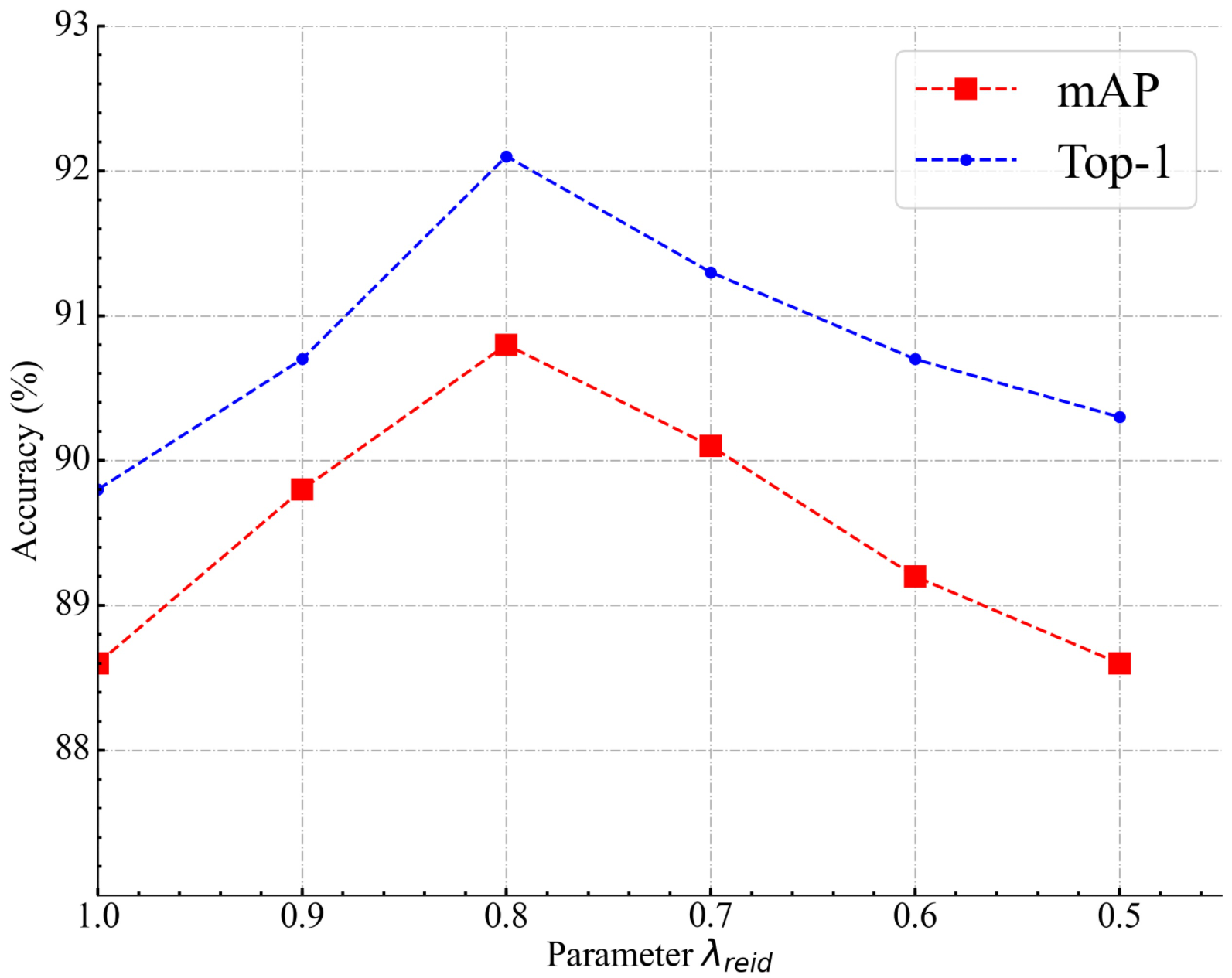}}
	\caption{The influence of different parameters on our method.}	
	\label{fig:CGC+UAM}	
	\vspace{0.1cm}
\end{figure}

\noindent \textbf{Evaluation of $\lambda_{sim}$.} Parameter $\lambda_{sim}$ is the trade-off coefficient between two kinds of similarity. Fig. \ref{components_a} shows that $\lambda_{sim} = 0.1 $ achieves the best performance. It is noteworthy that the parameter $\lambda_{sim}$ is different from parameters (\emph{e.g.}, the number of clusters or distance threshold) in previous clustering algorithm. $\lambda_{sim}$ stands for the trade-off between visual and context similarity, which is irrelevant to the datasets and thus works on different datasets.

\noindent \textbf{Evaluation of $\lambda_{reid}$.} Parameter $\lambda_{reid}$ is the balancing factor between paired cluster contrastive loss $L_{p}$ and unpaired cluster contrastive loss $L_{u}$. A large value of $\lambda_{reid}$ means a higher proportion of $L_{p}$ and $\lambda_{reid} = 1$ indexes the model does not leverage unpaired persons. Fig. \ref{components_b} shows that $\lambda_{reid} = 0.8 $ achieves the best performance. The performance drops when $\lambda_{reid}$ is less than 0.8, which indicates that the unpaired cluster contrastive loss $L_{u}$  plays an auxiliary role.

\subsection{More Implementation Details}
Limited by space, we present more implementation details in supplementary material to help others reproduce. We run all experiments on one NVIDIA Tesla V100 GPU.  Temperature hyper-parameter $\tau_c$ and momentum updating factor $m$ is set to 0.05 and 0.1, respectively. If not specify, $\lambda_{sim},\lambda_{reid}$ is set to 0.1 and 0.8 in manuscript experiments. The hard-sample mining scheme is followed as HHCL, and the number of instances is 4 and 16 for CUHK-SYSU and PRW respectively. Both detector and Re-ID model are optimized by the AdamW optimizer.

\end{document}